\documentclass[letterpaper, 10 pt, conference]{ieeeconf}  

\usepackage[T1]{fontenc}    
\usepackage[utf8]{inputenc} 

\usepackage{url}            
\usepackage{booktabs}       
\usepackage{amsfonts}       

\usepackage[dvipsnames]{xcolor} 

\usepackage{amsmath} 

\usepackage{amsthm}
\usepackage{amssymb} 
\usepackage{mathtools}
\usepackage{mathrsfs}
\usepackage{mathdots}
\usepackage[linesnumbered,ruled,vlined]{algorithm2e}

\usepackage{graphics} 
\usepackage{svg}
\usepackage{tikz}
\usetikzlibrary{shapes, shapes.geometric, arrows, calc}
\usetikzlibrary{bayesnet}
\usetikzlibrary{fadings}
\usetikzlibrary{patterns}
\usepackage{yhmath}
\usepackage{cancel}
\usepackage{gensymb}

\usepackage{pgfkeys}
\usepackage{standalone}

\usepackage{subcaption}
\usepackage{cite}
\usepackage{graphicx}
\usepackage{multirow}
\usepackage{threeparttable}
\usepackage{adjustbox}
\usepackage{aligned-overset}

\usepackage{float}
\usepackage{verbatim}
\let\labelindent\relax
\usepackage[shortlabels]{enumitem}

\IEEEoverridecommandlockouts                              

\title{\LARGE \bf A Solution to Adaptive Mobile Manipulator Throwing}

\author{Yang Liu \quad Aradhana Nayak \quad Aude Billard
\thanks{The authors are with Learning Algorithms and Systems Laboratory (LASA), EPFL, Switzerland. Email: \protect\url{{yang.liuu, aradhana.nayak, aude.billard}@epfl.ch}. Corresponding author: Yang Liu.
}
}

\begin{document}

\maketitle

\begin{abstract}
Mobile manipulator throwing is a promising method to increase the flexibility and efficiency of dynamic manipulation in factories. Its major challenge is to efficiently plan a feasible throw under a wide set of task specifications. We show that the mobile manipulator throwing problem can be simplified to a planar problem, hence greatly reducing the computational costs. Using machine learning approaches, we build a model of the object's inverted flying dynamics and the robot's kinematic feasibility, which enables throwing motion generation within 1 ms for given query of target position. Thanks to the computational efficiency of our method, we show that the system is adaptive under disturbance, via replanning on the fly for alternative solutions, instead of sticking to the original throwing plan. 
\end{abstract}

\section{Introduction}
Throwing, as one form of dynamic manipulation~\cite{mason1993dynamic, khurana2021learning}, can not only augment the feasible work space of the robot, but also increase the efficiency of object manipulation. This is highly desirable in applications such as logistics and handling of goods. However, automated robot manipulation with human-level speed is generally difficult to achieve and good solutions need to be hand-tuned for specific tasks~\cite{correll2016analysis}, which is a common drawback of model-based methods. On the other hand, with the emergence of machine learning, there is a growing popularity on using model-free methods that distill skills from offline training. With sufficient data coverage, the robot is able to handle a wide set of tasks once deployed. Despite impressive empirical result~\cite{zeng2020tossingbot}, model-free methods in the literature lead to algorithms which are difficult to generalize to new tasks and can not ensure successful task execution. 

\begin{figure}[t]
\centering
\includegraphics[scale=0.13]{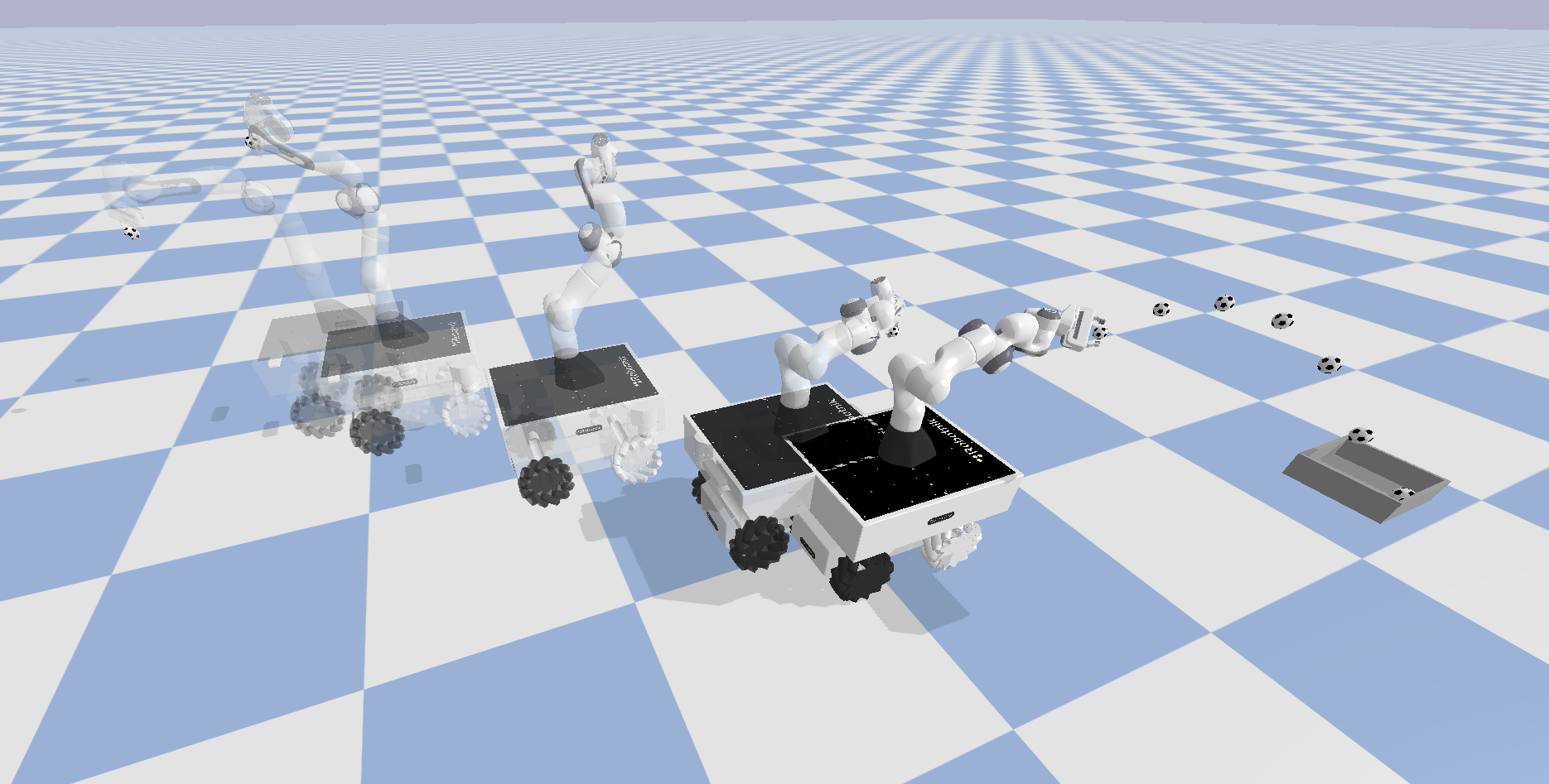}
\caption{Snapshots of a mobile manipulator throwing.}
\label{fig:snapshot-throw}
\end{figure}

We take a approach that combines the strength of model-based methods and model-free methods: model-free learning to model the complex non-linear object flying dynamics~\cite{kim2012estimating} and model-based approach to formulate throwing as a feasibility problem. To design an architecture that combines these two modules in harmony,
we start by asking the following questions:
\begin{itemize}
   \item What is an appropriate task representation for throwing?
   \item How to separate mundane computation \emph{offline} while ensuring reliable and efficient \emph{online} solution generation?
\end{itemize}
 To answer the above questions, we conduct an analysis on the structure of mobile manipulator-throwing problem, and design a data structure that enables efficient online query for throwing configurations. As a result, we propose a framework for mobile manipulator throwing that is able to generate throwing motion within 1 ms\footnote{The source code for our algorithms and simulations is available at:  \protect\url{https://github.com/epfl-lasa/mobile-throwing}}. We demonstrate a sample throw using a mobile manipulator in Fig.~\ref{fig:snapshot-throw}.
 
\subsection*{Related Work}
Previously, several approaches have been proposed to solve robot throwing problem~\cite{lynch1999dynamic, senoo2008high, mori20091, kober2012learning, gai2013motion, pekarovskiy2013optimal, taylor2019optimal, zeng2020tossingbot}. We notice that besides the dichotomy of model-based methods~\cite{lynch1999dynamic, senoo2008high, mori20091, gai2013motion, pekarovskiy2013optimal, taylor2019optimal} and model-free methods~\cite{ kober2012learning, zeng2020tossingbot} in the literature, there also exists a large difference on Operational Design Domain (ODD) among different works. We roughly categorize them into two types of ODDs: ad-hoc throwing with narrow ODD and generic throwing with wide ODD, see Table~\ref{tab:review}.

Most early works on robot throwing have been designed on a very narrow ODD with ad-hoc throwing solutions~\cite{lynch1999dynamic, senoo2008high, mori20091, gai2013motion, pekarovskiy2013optimal, miyashita2009control, kober2012learning}. They either worked on simple robots with low degrees of freedom~\cite{lynch1999dynamic, senoo2008high, mori20091, miyashita2009control, kober2012learning} or had a specific object to throw~\cite{gai2013motion, pekarovskiy2013optimal}. 

\begin{table}[t]
\centering
\caption{Classification of approaches on robot throwing.}
\begin{tabular}{l|c|c}
\hline
 & ad-hoc throwing & generic throwing \\ \hline 
 model-based & ~\cite{lynch1999dynamic, senoo2008high, mori20091, gai2013motion, pekarovskiy2013optimal, miyashita2009control} & 
\begin{tabular}{@{}c@{}}Optimization~\cite{lombai2009throwing, sintov2015stochastic}\\ Sampling~\cite{zhang2012sampling} \\ Ours \end{tabular} \\ \hline
model-free & ~\cite{kober2012learning} & End-to-end Learning~\cite{zeng2020tossingbot} \\ \hline
\end{tabular}
\label{tab:review}
\end{table}

Among the works aiming for generic throwing, many tried to accurately model the flying dynamics of objects as well as the robot dynamics and determine the optimal robot throwing trajectory via numerical optimization~\cite{lombai2009throwing,sintov2015stochastic}. However, as we shall see later in the paper, the optimization problem for throwing is hard to solve because of the nonconvex constraints. In \cite{sintov2015stochastic}, it takes $0.5$s to generate feasible throwing trajectory for a $3-$DOF robot. Therefore, it presumably needs at least several seconds for our mobile manipulator with $7-$DOF arm, which is not suitable for online motion generation.

Another approach for generic throwing is using sampling-based methods~\cite{zhang2012sampling} such as Probablistic Roadmap (PRM)~\cite{kavraki1996probabilistic} or Rapidly-exploring Random Trees (RRT)~\cite{kuffner2000rrt}. Although sampling-based methods provide probabilistic completeness, Zhang et al.~\cite{zhang2012sampling} reported that it takes up to 1 minute to get a throwing trajectory, which is far from human-level speed.


Recently, TossingBot~\cite{zeng2020tossingbot} offered an end-to-end learning approach for throwing. The throwing configuration is generated from throwing velocity predicted by trained model given throwing query and object image. The solution is generic and can handle a large set of objects and multiple target positions in the training set. However, the performance degrades for unseen query during training. While one could retrain the robot with new data, it is not clear how fast the robot can learn to throw new objects. 

\subsection*{Summary of Contribution}
Our approach complements the state-of-the-art by presenting a method for efficiently computing feasible throwing configurations and motions for mobile manipulators. In particular, we offer a comprehensive analysis on the problem of mobile manipulator throwing, including:
\begin{itemize}
    \item A geometric analysis, which exposes the independent variables and hence the task manifold of mobile manipulator throwing;
    \item A data structure for offline robot kinematic analysis that enables fast online query of throwing configurations.
\end{itemize}
\noindent With the above analysis, our framework enables the system to:
\begin{itemize}
    \item Throw a rigid-body object given its flying dynamics;
    \item Throw towards an arbitrary target position;
    \item Guarantee robot-feasibility and throwing-validity;
    \item Generate solutions within 1 ms and  achieve adaptive throwing.
\end{itemize}




\section{Method}
\begin{figure}[t]
\centering
\includegraphics[scale=0.15]{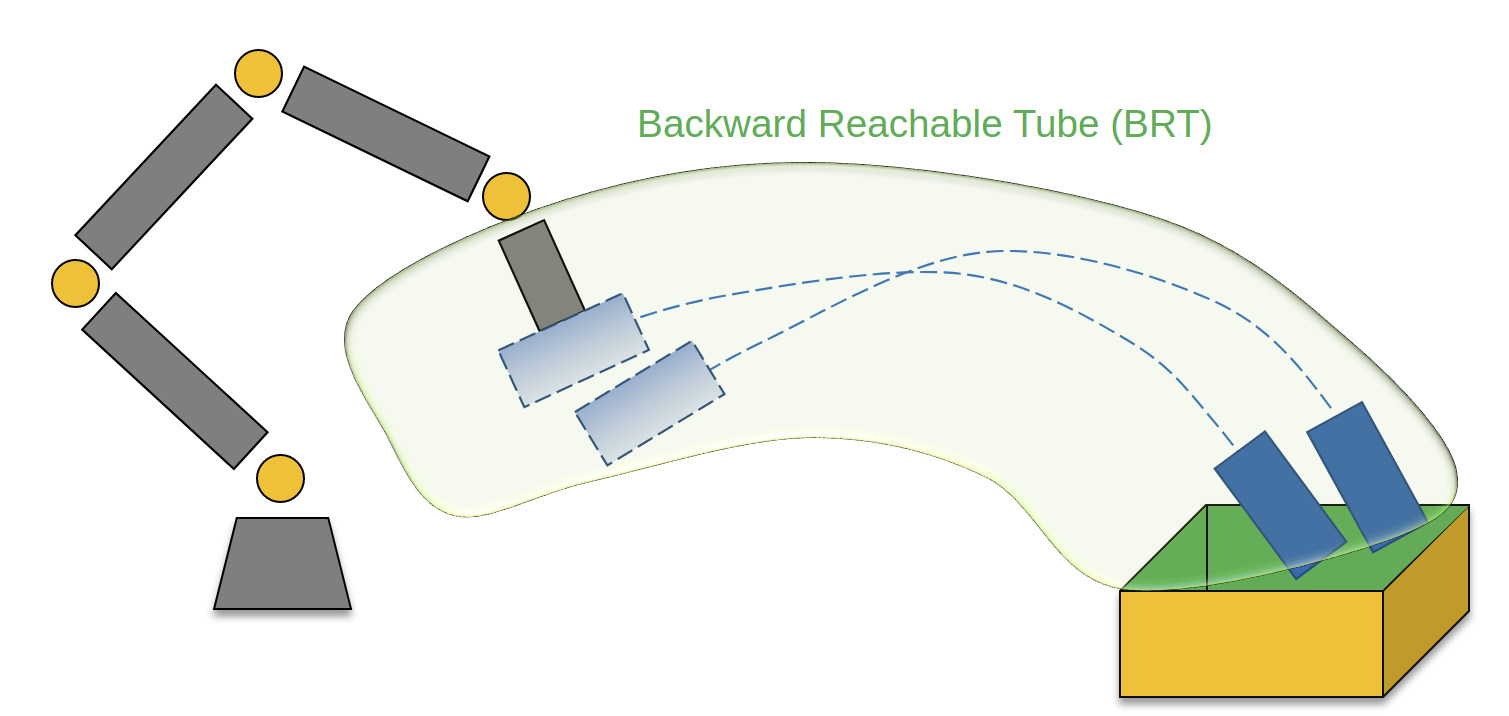}
\caption{The Backward Reachable Tube models the set of valid  throwing configurations, given object's flying dynamics. }
\label{fig:overall_idea}
\end{figure}
Given a throwing task specified by object flying dynamics, target landing position and possibly landing velocity range, there is a set of valid throwing configurations (throwing position and throwing velocity) varied in landing velocity and flying time. We call the set of valid throwing configurations Backward Reachable Tube (BRT). As shown in Fig.~\ref{fig:overall_idea}, our idea is to bring the object towards the BRT and release the object once the robot end-effector enters the BRT.

\subsection{Geometric modeling}
\begin{figure}[t]
    \centering
    \begin{subfigure}[b]{0.4\textwidth}
        \includegraphics[width=\textwidth]{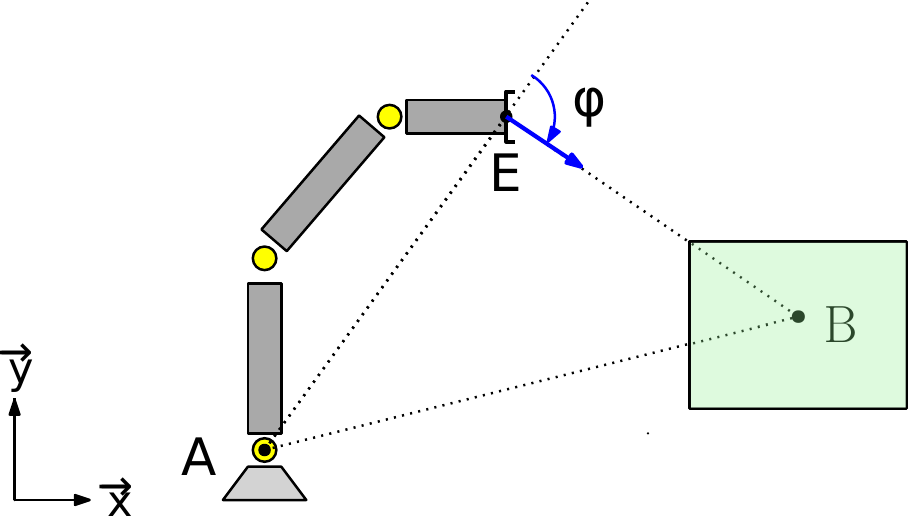}
        \caption{Top view (throwing triangle)}
        \label{fig:geometric-top}
    \end{subfigure}
    \begin{subfigure}[b]{0.4\textwidth}
        \includegraphics[width=\textwidth]{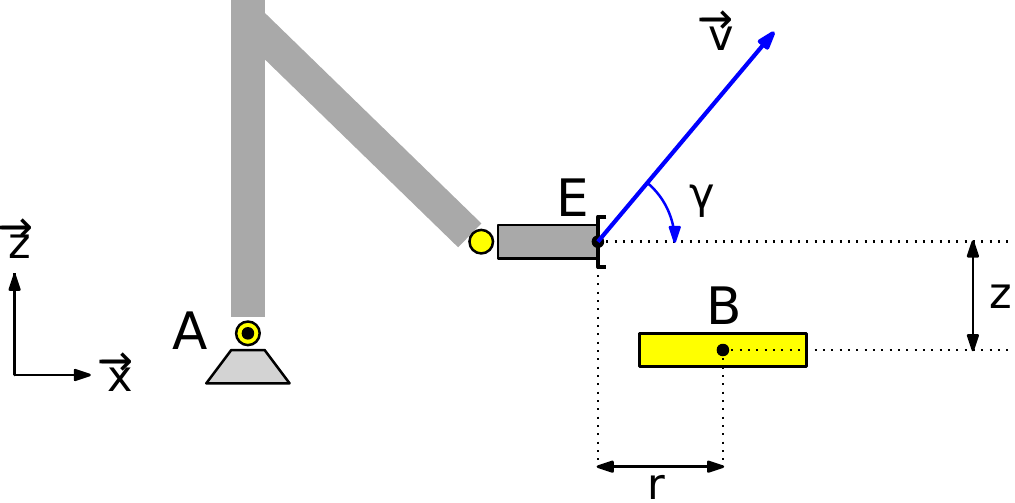}
        \caption{Side view (throwing plane)}
        \label{fig:geometric-side}
    \end{subfigure}
    \caption{Visualization of throwing triangle and throwing plane.}
\label{fig:geometry}
\end{figure}

\begin {table}[t]
\centering
\caption{Notations for geometric modeling in Fig.~\ref{fig:geometry}.}
\begin{tabular}{ l l }
\hline
 $A$ & Manipulator base \\ 
 $E$ & End effector \\  
 $B$ & Target box \\
 $\triangle AEB$ & Throwing triangle \\
 $\phi$ & Throwing yaw angle\\
 $r$ & Throwing range \\
 $z$ & Throwing height \\
 $\dot{r}$ & Horizontal throwing velocity \\
 $\dot{z}$ & Vertical throwing velocity \\
 $v$ & Throwing speed \\
 $\gamma$ & Throwing pitch angle \\ 
\hline
\end{tabular}
\label{tab:notation}
\end{table}
The geometric relationship between robot mobile manipulator and task throwing is shown in 
Fig. \ref{fig:geometric-top} and Fig. \ref{fig:geometric-side}. The notations are introduced in Table \ref{tab:notation}.


We observe that there exist geometric equivariance structures in mobile manipulator throwing, namely:
\begin{itemize}
    \item Mobile manipulator alone is equivariant in horizontal position because of the mobile base with omnidirectional wheels.
    \item Throwing alone is equivariant in incident direction.
    \item Mobile manipulator throwing can be described by throwing triangle $\triangle AEB$, which is equivariant in rotation around Z-axis at target box B. 
\end{itemize}

Above analysis showed that to find out the joint configuration for throwing, we only have to determine $\triangle AEB$. Furthermore, the triangle family around B could also be useful to avoid obstacle or communicate robot intent. These insights are crucial to our efficient and reliable solver.

\begin{figure}[t]
\centering
\hspace*{-0.5cm}\includegraphics[scale=0.6]{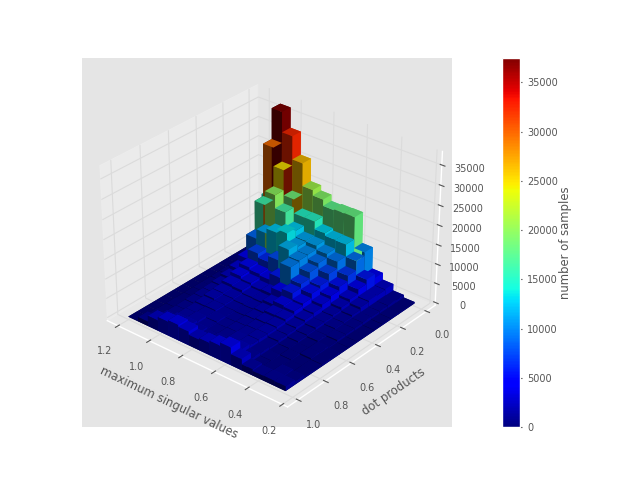}
\caption{Tradeoff between small yaw angle and large throwing velocity. For configurations with large feasible Cartesian velocity (large maximum singular value), the maximum velocities tend to be perpendicular to arm stretching direction (small dot product value). See Subsection~\ref{subsec:yaw} for detailed explanation.}
\label{fig:dot-product}
\end{figure}

\subsection{Consideration of yaw angle}
\label{subsec:yaw}
The essence of throwing is to move the object to somewhere not reachable by the robot end-effector. A typical choice in the literature~\cite{senoo2008high, lombai2009throwing, kober2012learning, zeng2020tossingbot} is to assume that the throwing yaw angle $\phi$ is zero, i.e., throwing velocity direction is aligned with end-effector horizontal direction $\overrightarrow{A E}_{xy}$ and hence the throwing triangle $\triangle AEB$ is degenerated to a line. However, we observe that throwing yaw angle $\phi = 0$ is not necessarily a good choice. Intuitively, if the arm is horizontally stretched, it is hard to generate velocity along the arm yet easy to generate velocity perpendicular to the arm. This hypothesis is validated in the following manipulabability analysis.

We sample 1 million joint configurations for a 7 DoF manipulator. For each sampled joint configuration $q$, we calculate their corresponding end-effector horizontal positions $\overrightarrow{A E}_{xy}(q)$ and Jacobian in $XY$ plane $J_{xy}(q)$. Then we group the 1 million data according to two attributes:
\begin{itemize}
    \item Maximum singular value of $J_{xy}$, denoted as $s_1(J_{xy})$, approximates the maximum horizontal velocity of the end-effector with unit joint velocity limit;
    \item The dot product between $\overrightarrow{A E}_{xy}$ and the singular vector associated with $s_1(J_{xy})$, approximates the ``alignment'' between maximum horizontal velocity direction and arm stretching direction $\overrightarrow{A E}_{xy}$, note that these two directions are aligned when the dot product is 1 and are perpendicular when the dot product is 0.
\end{itemize}

\noindent The resulting 3D histogram is shown in Fig.~\ref{fig:dot-product}.

One can observe that indeed there exists a tradeoff between small yaw angle and large throwing velocity, and if we predetermine $\phi=0$, a lot of appropriate throwing configurations are filtered out. Therefore, in this paper, we leave the throwing yaw angle as a decision variable and let the solver handle this tradeoff automatically.


\subsection{Backward Reachable Tube}\label{subsec:BRT}
In the throwing plane, we denote the object flying state as $x = [r, z, \dot{r}, \dot{z}]^\top$. The flying dynamics is described by a first order differential equation $\dot{x} = f(x), \, x \in \mathbb{R}^4$. We denote the flying trajectories of $f$ starting from state $x_0$ as $\zeta_{f, x_0}(t):[0, +\infty] \rightarrow \mathbb{R}^4$. We assume that a user is giving the robot a landing target set  $\mathcal{X}_l \subseteq \mathbb{R}^4$, describing the allowed landing position slack and allowed range of landing velocities. 

If there exist a position along the object flying trajectory that enters the landing target set, then any state on the trajectory segment before the entering state is a valid throwing configuration. Hence, if we aggregate all the trajectories that eventually enter the target landing set, we get the set of valid throwing configurations, which we call the Backward Reachable Tube (BRT). 

Mathematically, the BRT is defined as:
\begin{align*}
    BRT_{f, \mathcal{X}_l} = \{x_0 \, | \, \exists \, t \geq 0, \, \zeta_{f, x_0}(t) \in \mathcal{X}_l\}.
\end{align*}


The BRT can be generated from object flying dynamics as follows:

\subsubsection{Choose landing target set $\mathcal{X}_l$}
the landing target set can be simply specified as ``position box'' and ``velocity box''. Note that the horizontal velocity bound can be regarded as a limit to avoid flip out of the box and vertical velocity limit can be regarded as a limit to avoid object damage.

\subsubsection{Generating data inside BRT}
exact reachability analysis is generally computationally difficult for nonlinear dynamics. Our idea is to approximate BRT by sampling data inside it. Our observation is that flying dynamics is not memoryless and hence its corresponding Ordinary Differential Equation (ODE) can be solved backwards in time. Therefore, we sample landing configurations from landing target set $\mathcal{X}_l$, and solve the flying dynamics backwards in time as a Initial Value Problem (IVP) and then aggregate the ODE solutions. This is much more efficient than otherwise naive Boundary Value Problem (BVP) and hence can generate the BRT dataset quickly.

In this paper, the flying dynamics of an object is considered to be known apriori. This is assumed to be derived from state of the art system identification methods~\cite{schon2011system} or machine learning techniques~\cite{kim2012estimating}.

\subsubsection{Learning-based implicit representation of BRT}
with gathered data inside BRT, we use a Neural Network (NN) to represent the BRT as a level-set function $f_{BRT}(x): \mathbb{R}^4 \rightarrow \mathbb{R}$, where:
\begin{itemize}
    \item $f_{BRT}(x) > 0 \Leftrightarrow x \not \in BRT$;
    \item $f_{BRT}(x) \leq 0 \Leftrightarrow x \in BRT$;
    \item $f_{BRT}(x) = 0 \Leftrightarrow x \in \partial BRT$.
\end{itemize}

\subsection{Feasibility problem formulation}
The BRT only provides valid throwing configurations inside throwing plane in Euclidean space. To find throwing configurations in robot joint space, there are two types of nonlinear constraints to be considered:
\begin{itemize}
    \item Equality constraint: the throwing triangle $\triangle AEB$ defines a task manifold, i.e. for fixed $\overrightarrow{AB}$, the end-effector position $\overrightarrow{AE}$ must lie in the throwing plane defined by $\overrightarrow{EB}$, which imposes a nonlinear equality constraint in the joint configuration $q$.
    \item Inequality constraint: joint position limit and joint velocity limit.
\end{itemize}

These constraints can be expressed into the following Throwing Feasibility Problem (TFP):

\begin{align*}
\begin{gathered}
\text { Find } \left\{r, z, \dot{r}, \dot{z}, \overrightarrow{E B}, \overrightarrow{A B}, q, \dot{q} \right\}\\
\text { such that: }\left\{\begin{aligned}
r &= \left\|\overrightarrow{E B}_{x y}\right\| \\
z &=-\overrightarrow{EB}_{z} \\
\overrightarrow{EB} &=\overrightarrow{AB}-\overrightarrow{A E} \\
\overrightarrow{AE}&=fkine(q)\\
\vec{v}=\left[\begin{array}{l}
v_{x} \\
v_{y} \\
v_{z}
\end{array}\right] &=\left[\begin{array}{c}
\dot{r} \frac{\overrightarrow{E B}_{x y}}{\left\|\overrightarrow{E B}_{x y}\right\|} \\
\dot{z}
\end{array}\right] \\
\dot{q} &=J^{\dagger}(q) \vec{v} \\
q_{\min }  & \leq q  \leq q_{\max } \\
\dot{q}_{\min } & \leq \dot{q} \leq \dot{q}_{\max }\\
f_{BRT}&(r, z, \dot{r}, \dot{z})  \leq 0
\end{aligned}\right.
\end{gathered}\\
\tag{TFP}
\label{opt-full}
\end{align*}

\noindent where $fkine(\cdot)$ is the robot forward kinematics, and $\overrightarrow{EB}_{z}$ is projection of $\overrightarrow{EB}$ along vertical direction $Z$.
\ref{opt-full} is difficult to solve because of the nonlinear equality constraint. However, geometric analysis provides the following reasoning:
\begin{itemize}
    \item Once $(\overrightarrow{AB}, q)$ is fixed:
    \begin{itemize}
        \item $\triangle AEB$ is determined,
        \item $(\overrightarrow{EB}, r, z)$ is determined,
        \item $(\dot{r}, \dot{z})$ is to be determined;
    \end{itemize}
    \item Once $(\dot{r}, \dot{z})$ is further determined:
    \begin{itemize}
        \item $\dot{q} =J^{\dagger}(q) \vec{v}$;
    \end{itemize}
    \item Independent variables: $(\overrightarrow{AB}, q, \dot{r}, \dot{z})$.
\end{itemize}

From the above analysis, we expose the ``task manifold'' parameterized explicitly by $(\overrightarrow{AB}, q, \dot{r}, \dot{z})$ and hence eliminated the manifold constraint in the naive formulation. Now the problem reduces to the following formulation with only inequality constrains:
\begin{align*}
\begin{gathered}
\text { Find } \left\{\overrightarrow{A B}, q,\dot{r}, \dot{z}\right\}\\
\text { such that: }\
\left\{\begin{aligned}
& q_{\min }  \leq q  \leq q_{\max } \\
& \dot{q}_{\min } \leq  J^{\dagger}(q) \vec{v}(\overrightarrow{AB}, q, \dot{r}, \dot{z}) \leq \dot{q}_{\max }\\
& f_{BRT}(r(\overrightarrow{AB}, q), z(\overrightarrow{AB}, q), \dot{r}, \dot{z})  \leq 0
\end{aligned}\right.
\end{gathered}\\
\tag{TFP-Reduce}
\label{opt-reduce}
\end{align*}

\subsection{Robot kinematics analysis}\label{subsec:VelocityHedgehog}
The reduced feasibility problem still suffers from the nonconvex inequality constraint:
\begin{equation*}
    \dot{q}_{\min } \leq  J^{\dagger}(q) \vec{v}(\overrightarrow{AB}, q, \dot{r}, \dot{z}) \leq \dot{q}_{\max }
\end{equation*}

\noindent The constraint is linear in $(\overrightarrow{AB}, \dot{r}, \dot{z})$ but nonlinear in $q$, and it reflects different velocity capabilities at different arm configurations of the robot. As $q$ also determines end-effector position, this makes the robot kinematics analysis intertwined with throwing task. 

To mitigate the difficulty of finding a suitable $q$ to generate desired velocity at desired position, we propose \emph{velocity hedgehog}\footnote{The name comes from the spiky shape of the velocity distribution.}, a data structure that enables efficient online query of robot configurations. The velocity hedgehog discretizes the end-effector height $z$ and throwing direction $(\phi, \gamma)$ into cells. At each cell, it stores,

\begin{itemize}
    \item The maximum feasible end-effector speed $v_{max}$ at height $z$ along direction $(\phi, \gamma)$;
    \item The robot configuration $q$ that enables $v_{max}$.
\end{itemize}
Later, velocity hedgehog is used to generate batches of initial guesses for robot joint configurations in 1 millisecond. The algorithm to generate the velocity hedgehog is shown in Algorithm 1, where a Linear Program (LP) is solved as a subroutine to get maximum speed along direction $(\phi, \gamma)$ at robot configuration $q$ in the following manner:
\begin{align*}
\begin{gathered}
\max_{ s} \, s\\
\text { subject to: }\left\{\begin{aligned}
& \dot{q}_{\min} \leq  J^{\dagger}(q) \vec{v} \leq \dot{q}_{\max }\\
& \vec{v}
        =   s  \left[\begin{array}{l}
                \cos{\gamma}\cos{(\frac{\overrightarrow{AE}_y}{\overrightarrow{AE}_x}+\phi)} \\
                \cos{\gamma}\sin{(\frac{\overrightarrow{AE}_y}{\overrightarrow{AE}_x}+\phi)} \\
                \sin{\gamma}
                \end{array}\right]
\end{aligned}\right.
\end{gathered}\\
\tag{LP}
\end{align*}

\newcommand{\nextnr}{\stepcounter{AlgoLine}\ShowLn}
\makeatletter
\newcommand{\nosemic}{\renewcommand{\@endalgocfline}{\relax}}
\newcommand{\dosemic}{\renewcommand{\@endalgocfline}{\algocf@endline}}
\newcommand{\pushline}{\Indp}
\newcommand{\popline}{\Indm\dosemic}
\let\oldnl\nl
\newcommand{\nonl}{\renewcommand{\nl}{\let\nl\oldnl}}
\makeatother

\newcommand\mycommfont[1]{\footnotesize\ttfamily\textcolor{blue}{#1}}
\SetCommentSty{mycommfont}
\SetKwInOut{Input}{Input}
\SetKwInOut{Output}{Output\,}
\SetKwInOut{Data}{Data}
\SetKwProg{Tree}{Tree}{}{EndTree}

\begin{algorithm}[!htp]
  \SetAlgoLined
  \caption{Algorithm to get velocity hedgehog}
  \Input{robot model with forward kinematics and differential forward kinematics $robot$}
  \Output{max\_z\_phi\_gamma, q\_z\_phi\_gamma}
  \Data{robot joint position limit $(q_{min}, q_{max})$,\\ 
        robot joint velocity limit $(\dot{q}_{min}, \dot{q}_{max})$,\\
        joint grid size $\Delta q$, \\ 
        velocity hedgehog grids $Z, \Phi, \Gamma$\linebreak}

\tcc{Build robot dataset $\mathcal{X}$}
$\mathcal{Q}  \gets ComputeMesh(q_{min}, q_{max}, \Delta q)$ \\
\tcc{Filter out $q$ with small singular value}
$\mathcal{X} \gets FilterBySinguarValue(\mathcal{Q})$ \\

\tcc{Group data by $Z$}
$\{\mathcal{X}_z\} \gets GroupBy(\mathcal{X}, Z)$\\
\tcc{Initialize velocity hedgehog}
$\text{max\_z\_phi\_gamma} \gets zeros([\#Z, \#\Phi, \#\Gamma])$\\
$\text{q\_z\_phi\_gamma} \gets arrays([\#Z, \#\Phi, \#\Gamma, \#joints])$\\
\tcc{Build velocity hedgehog}
\For{$[z,\phi,\gamma, data ] \in Z \times \Phi\times \Gamma \times \mathcal{X}_z$}
               { \tcc{Get max. speed along ($\phi, \gamma$) at joint configuration $q$}
                $res \gets LP(\phi, \gamma, data.q, robot, \dot{q}_{min}, \dot{q}_{max})$\\
                \If {$res > \text{max\_z\_phi\_gamma}(z, \phi, \gamma)$}{
                    $\text{max\_z\_phi\_gamma}(z, \phi, \gamma) \gets res$\\
                    $\text{q\_z\_phi\_gamma}(z, \phi, \gamma) \gets data.q$
                }
    }
\Return $\text{max\_z\_phi\_gamma}, \text{q\_z\_phi\_gamma}$
\label{algo:hedgehog}
\end{algorithm}

  

\begin{algorithm}
  \SetAlgoLined
  \caption{Algorithm to get initial guesses}
  \Input{robot velocity hedgehog, BRT dataset}
  \Output{Set of initial guess for joint configuration and throwing configuration $(q, \phi, x)$}
  
  $\mathcal{X}_{z,\gamma} \gets GroupBy(\mathcal{X}_{BRT}, Z, \Gamma)$\\
  $q\_guess, \phi\_guess, x\_guess \gets empty$
  
  $idxs \gets where(max\_z\_phi\_gamma > \mathcal{X}_{z,\gamma})$\\
  $q\_guess = q\_z\_phi\_gamma[idxs]$\\
  $\phi\_guess = \Phi[idxs]$\\
  $x\_candidates = \mathcal{X}_{z,\gamma}[idxs]$\\
  \Return $q\_guess, \phi\_guess, x\_guess$
  
\label{algo:initial-guess}
\end{algorithm}

A typical velocity hedgehog is shown in Fig. \ref{fig:robot_hh_0}, indicating that the feasible velocity set is not convex but appears as needles with different lengths along different directions and resembles a hedgehog.

\definecolor{pptgreen}{RGB}{0,176,80}
\definecolor{pptblue}{RGB}{0,112,197}
\begin{figure}[t]
    \centering
    \includegraphics[scale=0.35]{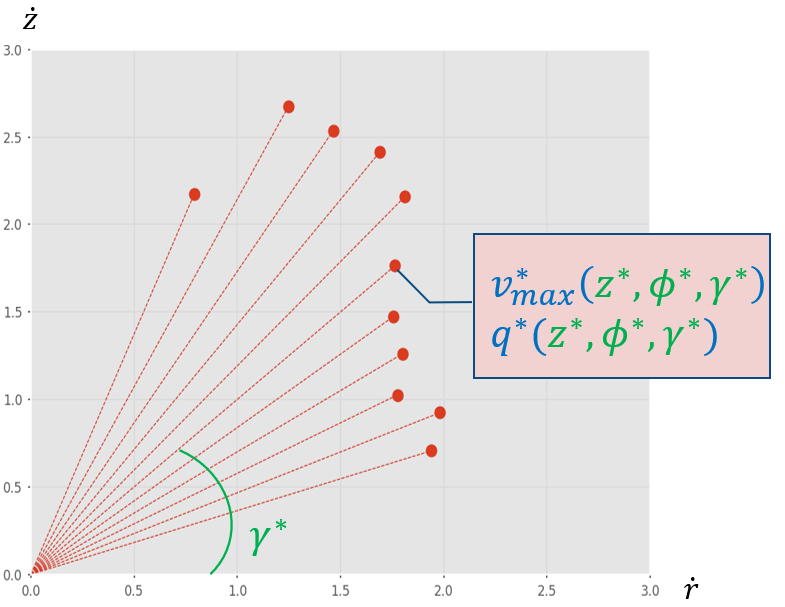}
    \caption{Illustration of a typical velocity hedgehog. Red needles represents the feasible velocities along different throwing pitch angles with fixed throwing height $z$ and throwing yaw angle $\phi$, which are stored in velocity hedgehog. Given certain throwing height and throwing direction \textcolor{pptgreen}{$\pmb{(z^*, \phi^*, \gamma^*)}$} query, velocity hedgehog gives maximum feasible velocity \textcolor{pptblue}{$\pmb{v^*_{max}}$} and the joint configuration  \textcolor{pptblue}{$\pmb{q^*}$} that enables \textcolor{pptblue}{$\pmb{v^*_{max}}$}.}
    \label{fig:robot_hh_0}
\end{figure}

\begin{figure*}[t]
\centering
\includegraphics[width=20cm,height=8cm,keepaspectratio]{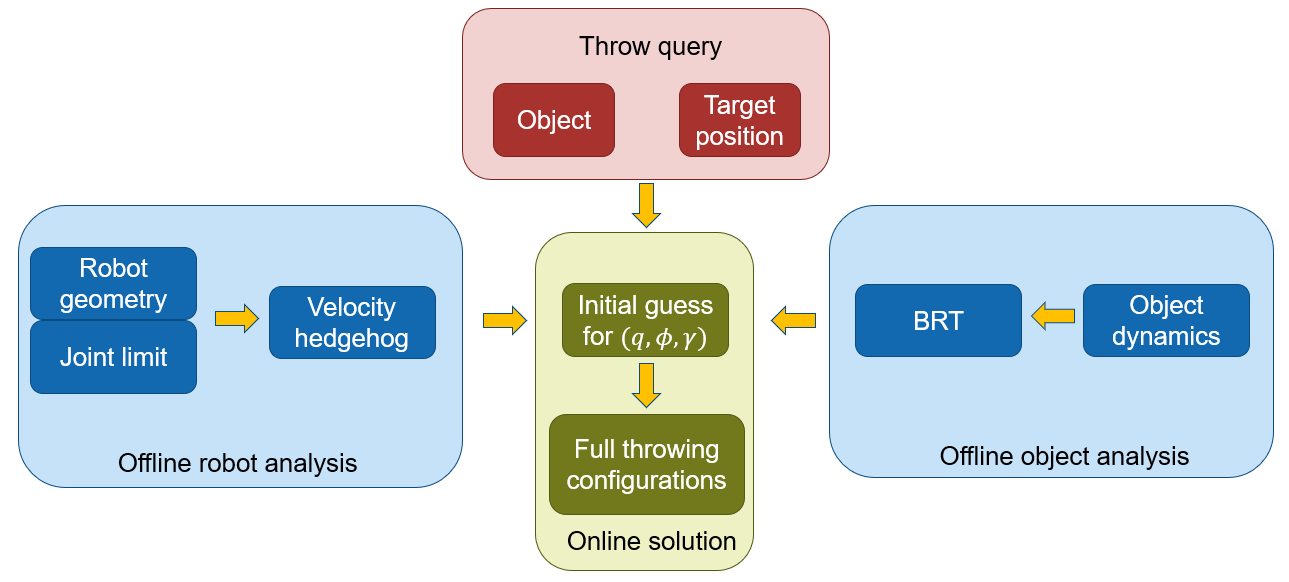}
\caption{Framework to obtain throwing configurations. The velocity hedgehog and BRT are obtained as described in Sections \ref{subsec:VelocityHedgehog} and \ref{subsec:BRT}, respectively. The initial guess is obtained by combining them as outlined in Section \ref{subsec:Combined}.}
\label{fig:architechture}.
\end{figure*}

\subsection{Combining robot velocity hedgehog with BRT to determine throwing configurations}\label{subsec:Combined}
Here we introduce a simple method to combine robot velocity hedgehog and BRT, resulting in an efficient method to get qualitatively different initial guesses for $(q, \phi, x)$ quickly. We first group the BRT data according to height $z$ and throwing pitch angle $\gamma$. Then for BRT data in bin indexed by $\hat{z}$ and $\hat{\gamma}$, the initial guesses for throwing state are the BRT data whose flying speed is smaller than the maximum feasible velocity at a certain $(\hat{z},\hat{\gamma})$ stored in velocity hedgehog. From this matching operation, we can also read out the corresponding joint configurations and throwing yaw angles from velocity hedgehog, resulting in initial guesses of $(q, \phi, x)$. The detailed algorithm is shown in Algorithm 2. The obtained feasible initial guesses can be fed into Problem~\ref{opt-reduce} to be further refined. With $(q, \phi, x)$ determined, all the other decision variables can be written in closed-form as shown in Problem~\ref{opt-full}. As a result, the overall architechture to solve~\ref{opt-full} efficiently and reliably is shown in Fig.~\ref{fig:architechture}.


  


\section{Experiment}
We verify our method in the simulated environment PyBullet~\cite{coumans2021} with following setups:
\begin{itemize}
    \item Known accurate flying dynamics;
    \item Perfect grasping (object won't fall off while on the way to throwing configuration);
    \item Instant gripper opening;
    \item Perfect throwing trajectory tracking;
    \item Disabled collision checking between robot and box.
\end{itemize}
The object is a ball with radius $5$ cm and the target box dimensions are $5$ cm $\times$ $25$ cm $\times$ $25$ cm.
\noindent This paper's emphasis is on finding valid joint space throwing configuration in a reliable and computational efficient manner.
Therefore, the above assumptions are mild.

\subsection{BRT data generation}
\label{subsec:brt-data-exp}
We describe the data generation procedure for object flying states inside BRT as defined in Subsection~\ref{subsec:BRT}. For the sake of simplicity, we use pure projectile flying dynamics with gravity $g = 9.81 \text{ m} \slash \text{s}^2$, i.e.:

\begin{align*}
    \dot{x} = f(x) =\frac{d}{dt}\left[\begin{array}{l}
                    r \\
                    z \\
                    \dot{r} \\
                    \dot{z}
                    \end{array}\right] = 
                    \left[\begin{array}{l}
                    \dot{r} \\
                    \dot{z} \\
                    0 \\
                    -g
                    \end{array}\right],
\end{align*}

with following target set $\mathcal{X}_l$: 

\begin{align*}
    \mathcal{X}_l = \left\{
    \begin{array}{c}
                    r = 0\\
                    z = 0\\
                    0.2 \leq \dot{r} \leq 2.0\\
                    -5.0 \leq \dot{z} \leq -2.0
                    \end{array}
    \right\}.
\end{align*}

To approximate the BRT, we sample 2160 landing states inside $\mathcal{X}_l$, solve the flying dynamics backwards in time for $1$ second to get the 2160 flying trajectories, then aggregate all the data points on the trajectories, filter out the data with high velocities that is for sure not feasible be the robot ($|\dot{r}| > 5.0$, $|\dot{z}| > 5.0$), finally yield 75000 throwing configurations in throwing plane coordinates.

\subsection{Neural network-based BRT representation}
\label{subsec:brt-nn-exp}
\begin{figure}[t]
\centering
\includegraphics[scale=0.55]{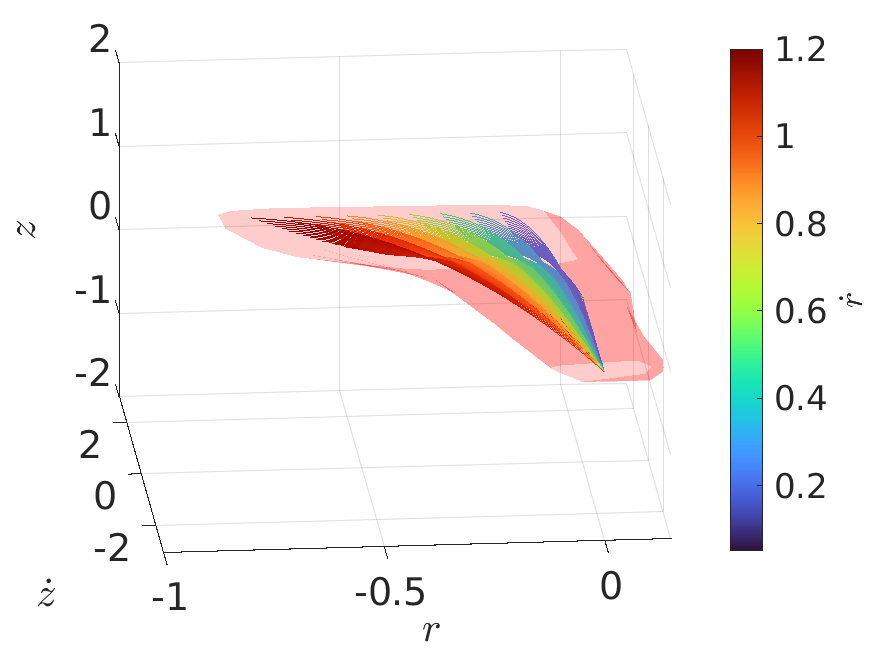}
\caption{Object flying trajectory family (rainbow) and boundary of learned neural network-based BRT representation (pink). Decision boundary strictly encompasses the flying trajectories because neural network was trained with data augmented in $r$ and $z$ directions.}
\label{fig:brt_nn}
\end{figure}

From the pipeline of BRT data generation, we also obtain 80000 throwing configurations outside BRT via sample 2160 landing configurations outside target set $\mathcal{X}_l$. These is treated as negative data, together with the above 75000 positive data as the dataset to train a neural network-based BRT representation. To make sure the BRT is a dense 4D volume, we augment the dataset by copying and shifting the data in $r$ and $z$ directions. We split the dataset to $70\%$ training data and $30\%$ testing data. The neural network contains 4 layers with size 4-64-64-2 and sigmoid activation function. We use Adam optimizer for training, and the training takes 2 minutes for 10 epochs. The trained neural network achieved $97\%$ classification accuracy on the test set. 
The decision boundary $f_{BRT}(x) = 0$ of the neural network is illustrated in Fig.~\ref{fig:brt_nn}. 

\subsection{Robot velocity hedgehog generation}
\label{subsec:velo-hedge-exp}
We denote a scalar array consists of values from $a$ to $b$ with equal interval $c$ as $[a:c:b]$. We discretize the throwing height with grid $Z=[0:0.05:1.2]$, the throwing yaw angle with grid  $\Phi=[-90:15:90]$ and the throwing pitch angle with grid $\Gamma=[20:5:70]$, yield 3289 cells of robot throwing candidates. The robot velocity hedgehog is generated using Algorithm 1 with 1 million joint state samples. 

\subsection{Trajectory generation towards throwing configuration}
\label{subsec:traj-exp}
Given initial state of the robot $(q_0, \dot{q}_0)$ and feasible throwing configuration $(q_d, \dot{q}_d)$, we use Ruckig~\cite{berscheid2021jerk} to generate jerk-limited time-optimal trajectory to move the robot towards throwing configuration. Then the robot opens the gripper to release the object at the end of the trajectory.


\section{Evaluation and discussion}

\subsection{Computational Efficiency}

Here we summarize the computation time of offline stage in Table~\ref{tab:time-offline} and online stage in Table~\ref{tab:time-online}.

\begin{table}[t]
\centering
\caption{Offline computation time.}
\begin{tabular}{lc}
\toprule
Algorithm stage & Running time\\ 
\midrule
Velocity hedgehog generation & 7 hours\\
BRT data generation & 1 second\\
Training BRT neural network & 2 minutes\\
\bottomrule
\end{tabular}

\label{tab:time-offline}
\end{table}

Although the algorithm takes a long time to generate velocity hedgehog, it only need to run once for every robot with the same geometry. In terms of new object, given its flying dynamics, the algorithm is efficient (2 minutes) to be ready to throw.

\begin{table}[t]
\centering
\caption{Online computation time.}
\begin{tabular}{lc}
\toprule
Algorithm stage & Running time per solution \\ 
\midrule
Initial guess & 20 $\mu s$\\
Full throwing configuration & 250 $\mu s$\\
Trajectory generation & 200 $\mu s$\\
\\
Overall & 500 $\mu s$\\
\bottomrule
\end{tabular}
\label{tab:time-online}
\end{table}

\begin{table}[t]
\centering
\caption{Solver feasibility.}
\begin{tabular}{lcccc}
\toprule
Target height & $z=-0.2$ & $z=0.0$ & $z=0.2$ & $z=0.5$ \\ 
\midrule
Number of solutions & 11955 & 10504 & 7118 & 2422\\
\bottomrule
\end{tabular}
\label{tab: solver-feasibility}
\end{table}

\begin{figure}[t]
\centering
\includegraphics[scale=0.13]{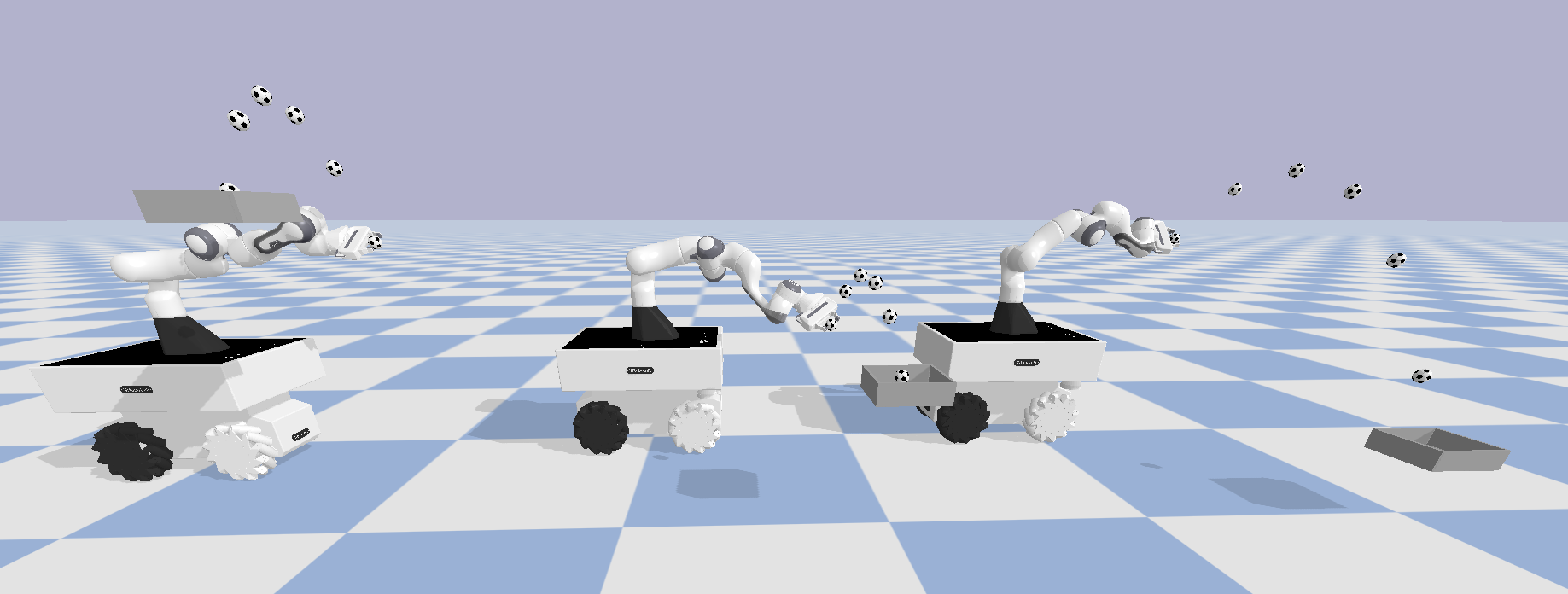}
\caption{Throw to three different target heights.}
\label{fig:diff-zs}
\end{figure}

\begin{figure}[t]
\centering
\includegraphics[scale=0.21]{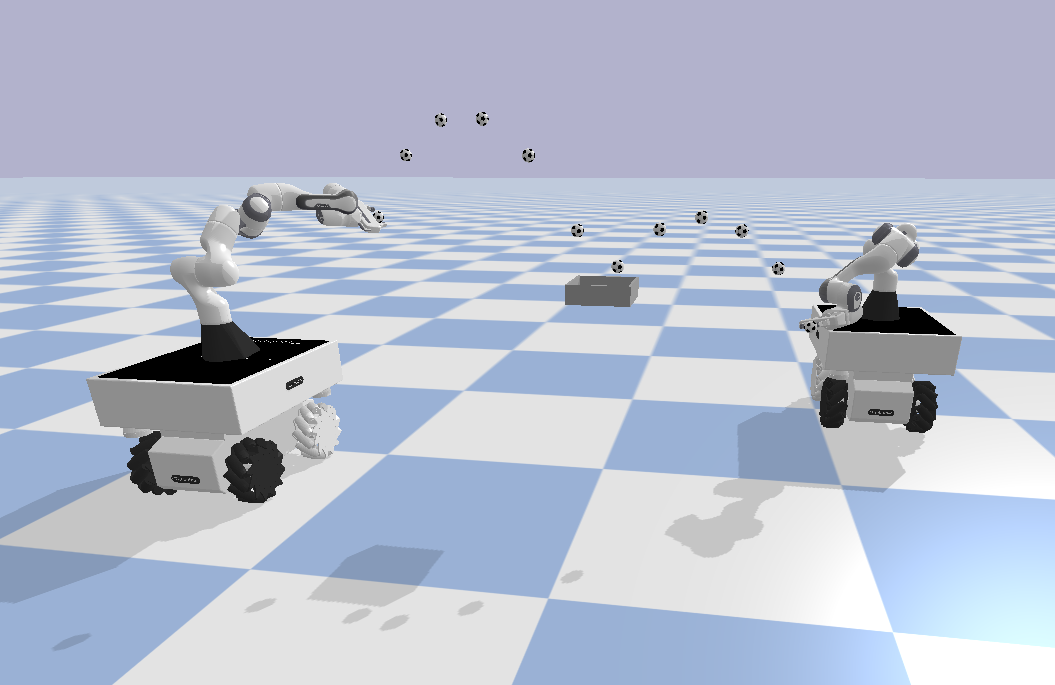}
\caption{Two qualitatively different throwing configurations, note the different heights of the end-effectors.}
\label{fig:quality-different}
\end{figure}

With our proper separation of offline computation and online computation, the algorithm is quick to generate throwing motion given throwing query.

\subsection{Reliability of solver}
Our throwing geometry analysis and robot velocity analysis results in highly reliable solver, which provides batches of throwing configurations. In comparison, generic solvers have a hard time to find one feasible configuration, and they have to be carefully tuned and be provided with good initial guess by human. Table~\ref{tab: solver-feasibility} demonstrates that our solver provides ample of throwing configurations for various target heights. In Fig.~\ref{fig:diff-zs}, we show three throws towards targets with different heights. Furthermore, for the same target box, the solver is able to find out qualitatively different solutions, as shown in Fig.~\ref{fig:quality-different}.

\subsection{Throw success rate in simulation}
We systematically assessed our algorithm's ability to generate a valid throw for a set of target height queries in the range of $Z = [-1.2:0.1:0.9]$. The algorithm generated in total 10366 throwing trajectories from the same initial configuration of the robot. We simulated the generated throwing trajectories for validation. As a result, 10306 $(99.4\%)$ throws were successful throw. As we observe the failure in $0.6\%$ of the cases, they are all due to collision: ball collided either with robot or box while flying. Considering the fact that we do not handle collision avoidance explicitly, this is anticipated.

\subsection{On throwing configuration selection}
The proposed algorithm outputs batches of throwing configurations, which are valid (object will fly into the target box) and feasible (throwing state satisfies robot joint position and velocity constraints). However, at this moment, we haven't determine the final throwing configuration from the solution set, which is to be executed by the robot. Selecting throwing configuration should consider broader context or higher level specifications of the use cases. For example, if one seeks for the time-optimal solution, the best throwing configuration would depend on the initial configuration of the robot. If one seeks for a solution with lower risk, the throwing configuration selection procedure would bias towards small-range throws. While here in this paper, we emphasize on the guaranteed validity and feasibility of the solver, throwing configuration selection is out of the scope of this work.

\subsection{Adaptive throwing}
\begin{figure}[t]
\centering
\includegraphics[scale=0.22]{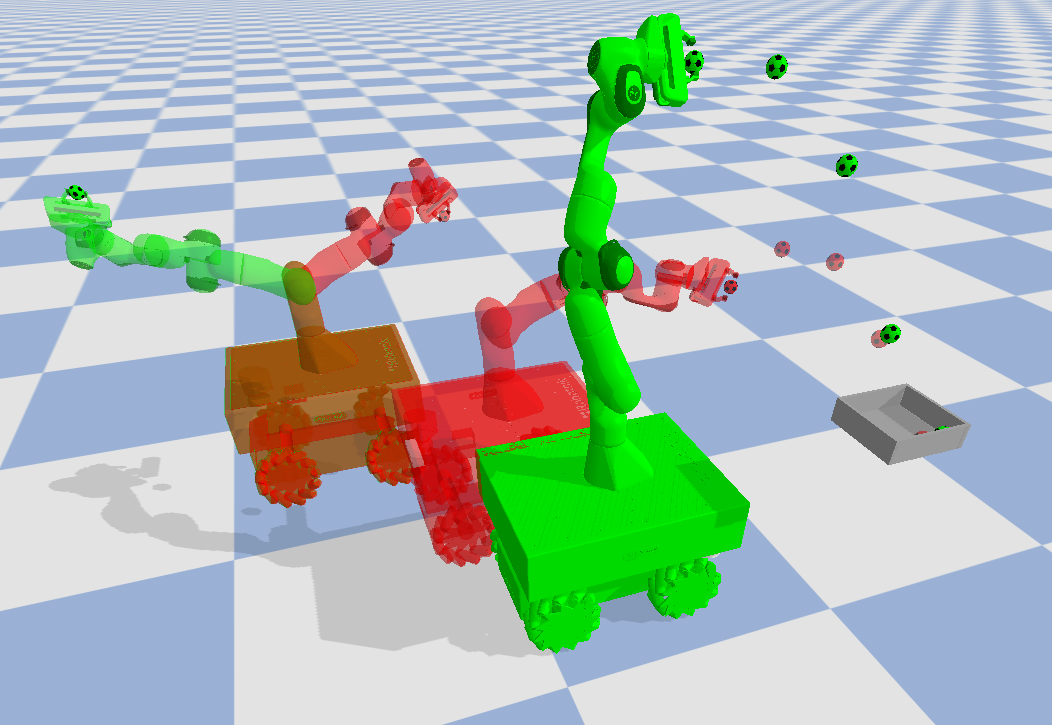}
\caption{Adaptive throw as the reaction to disturbance. See Subsection~\ref{subsec:adaptive} for detailed explanation.}
\label{fig:adaptive-throw}
\end{figure}

\
\label{subsec:adaptive}
Our method also enables adaptive throwing. As shown in Fig.~\ref{fig:adaptive-throw}, while the robot was moving towards the originally planed time-optimal throwing configuration (\textcolor{red}{red right}), it suddenly got disturbed to a different configuration during execution (from \textcolor{red}{red left} to \textcolor{green}{green left}). There are two strategies to handle this disturbance:
\begin{enumerate}[(a)] 
    \item Recalculate the trajectory from the disturbed configuration to the original throwing configuration;
    \item Discard the originally planned throwing configuration, sample 100 initial guesses obtained before execution from Algorithm 2, find a new throwing configuration that is closest in time to the disturbed configuration.
\end{enumerate}

We summarize the computation time and duration of obtained throwing trajectory in Table~\ref{tab:time-disturb}. As shown in Table~\ref{tab:time-disturb}, strategy (b) trades 15 ms computation time for 670 ms task execution time, which is preferable. As a result, the robot switched to another valid throwing configuration (\textcolor{green}{green right}) that is actually closer to the disturbed configuration.

\begin{table}[t]
\centering
\caption{Computation time of different strategies to handle disturbance.}
\begin{tabular}{cccc}
\toprule

Strategy & Computation time & Trajectory duration & Total\\ 
\midrule
(a) & 0.2 ms & 1800 ms & $\sim$1800 ms\\
(b) & 15 ms & 1134 ms & $\sim$1150 ms\\
\bottomrule
\end{tabular}
\label{tab:time-disturb}
\end{table}

\section{Conclusion}
In this paper, we propose a solution to mobile manipulator throwing. The merits of our method are:
\begin{itemize}
    \item Ability to handle arbitrary rigid object given its flying dynamics;
    \item Ability to handle arbitrary target position within the robot kinematic limits;
    \item Capacity for adaptive throwing.
\end{itemize}

We validate our framework in simulation and the algorithm is shown to be efficient and reliable. In the future, we will implement our algorithm on the real robot platform to test its effectiveness. Moreover, we plan to investigate the influence of modeling error in flying dynamics and of trajectory tracking accuracy on the real throwing set-up.


\section*{ACKNOWLEDGMENT}
\noindent The authors acknowledge the H2020 ICT-46-2020 EU project DARKO for supporting this work. 
\bibliographystyle{IEEEtran}
\bibliography{references.bib}
\end{document}